\documentclass[onecolumn]{IEEEtran}

\usepackage{balance}
\usepackage{amsmath}
\usepackage{graphicx}
\usepackage{booktabs}
\usepackage{url}

\newcommand{\best}[1]{\textbf{#1}}

\title{Transfer Learning for Covert Speech Classification Using EEG Hilbert Envelope and Temporal Fine Structure}

\author{
    \IEEEauthorblockN{
      Saravanakumar Duraisamy\IEEEauthorrefmark{1} \quad
      Mateusz Dubiel\IEEEauthorrefmark{1} \quad 
      Maurice Rekrut\IEEEauthorrefmark{2} \quad 
      Luis A. Leiva\IEEEauthorrefmark{1}
    }\\
    \IEEEauthorblockA{\IEEEauthorrefmark{1}University of Luxembourg, Luxembourg}\\
    \IEEEauthorblockA{\IEEEauthorrefmark{2}German Research Center for Artificial Intelligence (DFKI), Germany}
    \thanks{To appear in the 50th Proceedings of the IEEE International Conference on Acoustics, Speech, and Signal Processing (ICASSP), 2025.}
}

\begin{document}

\maketitle

\begin{abstract}
Brain-Computer Interfaces (BCIs) can decode imagined speech from neural activity. However, these systems typically require extensive training sessions where participants imaginedly repeat words, leading to mental fatigue and difficulties identifying the onset of words, especially when imagining sequences of words. This paper addresses these challenges by transferring a classifier trained in overt speech data to covert speech classification. We used electroencephalogram (EEG) features derived from the Hilbert envelope and temporal fine structure, and used them to train a bidirectional long-short-term memory (BiLSTM) model for classification. Our method reduces the burden of extensive training and achieves state-of-the-art classification accuracy: 86.44\% for overt speech and 79.82\% for covert speech using the overt speech classifier.
\end{abstract}


\section{Introduction}
\label{sec:intro}

Covert (or silent) speech refers to speech that is produced without any audible sound~\cite{freitas2017introduction,denby2010silent}. 
Various approaches have been proposed to measure covert speech, 
including electrical activity of articulatory muscles
or ultrasound-based lipreading~\cite{kapur2018alterego,kimura2019sottovoce}. 
A more natural approach is to measure the neural correlates of speech directly from brain activity, 
using Brain-Computer Interfacing (BCI) technology 
in the so-called speech imagery or imaginary speech BCIs~\cite{wang2013analysis,gonzalez2020silent,brumberg2010brain}. 

Since speech is generated in the brain before it is articulated~\cite{brumberg2016spatio}, 
decoding it directly from brain signals bypasses the need for muscle movements. 
This makes the process silent and imperceptible to others. 
In essence, imagined speech BCIs aim to interpret words or phonemes from brain activity alone, 
without any audible sound or physical movement involved.
Although invasive methods such as electrocorticography (ECoG) have yielded promising results~\cite{proix2022imagined,luo2023brain}, noninvasive approaches such as electroencephalography (EEG) 
have also shown potential in decoding imagined speech~\cite{lopez2022state,reddy2024multivariate}. 
However, these results are often confined to controlled lab environments and require extensive and exhausting training sessions.

A major challenge in imagined speech BCIs is the tedious and error-prone training process, in which participants silently repeat words, leading to potential data mislabeling
due to unverified output and variability in speech onset, offset, and speed. 
We propose a transfer learning approach to address these challenges
by training a classifier on EEG data recorded during overt speech
and applying it to EEG data of the same person silently repeating the words. 
This method not only reduces the mental and physical demands of training,
but also enhances productivity by allowing simultaneous interaction and classifier training.

We used the data recorded during a previous study~\cite{rekrut2022improving} in which participants navigated a virtual robot through a maze 
using both overt and imagined speech,\footnote{Dataset available at \url{https://doi.org/10.5281/zenodo.14645653}} 
to apply a set of feature extraction methods (described later) 
and transferred a classifier from overt to imagined speech data, 
drawing inspiration from related work, which we discuss in the next section.
Our research shows promise for developing BCI systems that can accurately decode covert speech using brain signals. 
This could enable truly usable applications in speech communication 
for people with disabilities or new communication tools.

\section{Related work}
\label{sec:relatedwork}

Garcia-Salinas et al.~\cite{garcia2019transfer} recorded EEG data from 27 participants
while they silently spoke five words (\textsc{Select}, \textsc{Right}, \textsc{Left}, \textsc{Down}, and \textsc{Up}). 
A Naive Bayes classifier trained on a subset of words and tested on all five achieved 68.93\% accuracy.
Lee et al.~\cite{lee2020eeg} investigated common spatial and temporal features 
between overt and imagined speech from seven participants and 12 words, 
achieving a 59.9\% accuracy for overt speech and 16.2\% for imagined speech, both above chance levels. 
They suggested potential for imagined speech classification based on overt speech features 
but did not attempt a direct transfer. 
Watanabe et al.~\cite{watanabe2020synchronization} examined whether 
EEG acquired during speech perception and imagination shared a signature envelope with EEG from overt speech. 
Their study, involving 18 participants and three words, showed that classifiers trained on imagined speech EEG envelopes 
could achieve 38.5\% accuracy when tested on overt speech envelopes. 
These findings reinforce the similarity between the two paradigms.

Komeiji et al.~\cite{komeiji2024feasibility} developed a Transformer-based model 
that decoded sentences from covert speech using ECoG signals. 
They outperformed a BiLSTM model in both overt and covert speech tasks. 
Interestingly, training the model on overt speech achieved similar performance for decoding covert speech 
compared to training on covert speech itself. 
This is a significant finding, as it bypasses the difficulty of collecting covert speech data. 
Surprisingly, to the best of our knowledge, no research has yet attempted 
a direct transfer of a classifier from overt to imagined speech. 
Our work therefore addresses this longstanding research gap, 
offering a more productive training process while maintaining high accuracy.

\section{Methodology}
\label{sec:Methodology}

Our goal is to apply transfer learning by using a model trained on overtly spoken EEG data 
and transferring it to classify covert (inner/imagined) speech EEG data. 
We used the dataset of Rekrut et al.~\cite{rekrut2022improving} 
which acquired overt and covert speech EEG data with a game-like setup as described below.

\subsection{Experimental design and procedure}

\subsubsection{Participants}
Fifteen healthy, right-handed subjects (11 male, 4 female, average age 26.8 years) participated in the study. 
All participants were fluent in English. 
The study was approved by the Ethical Review Board 
of the Faculty of Mathematics and Computer Science at Saarland University.

\subsubsection{Design} 
\begin{figure}[!ht]
\centerline{\includegraphics[width=\linewidth]{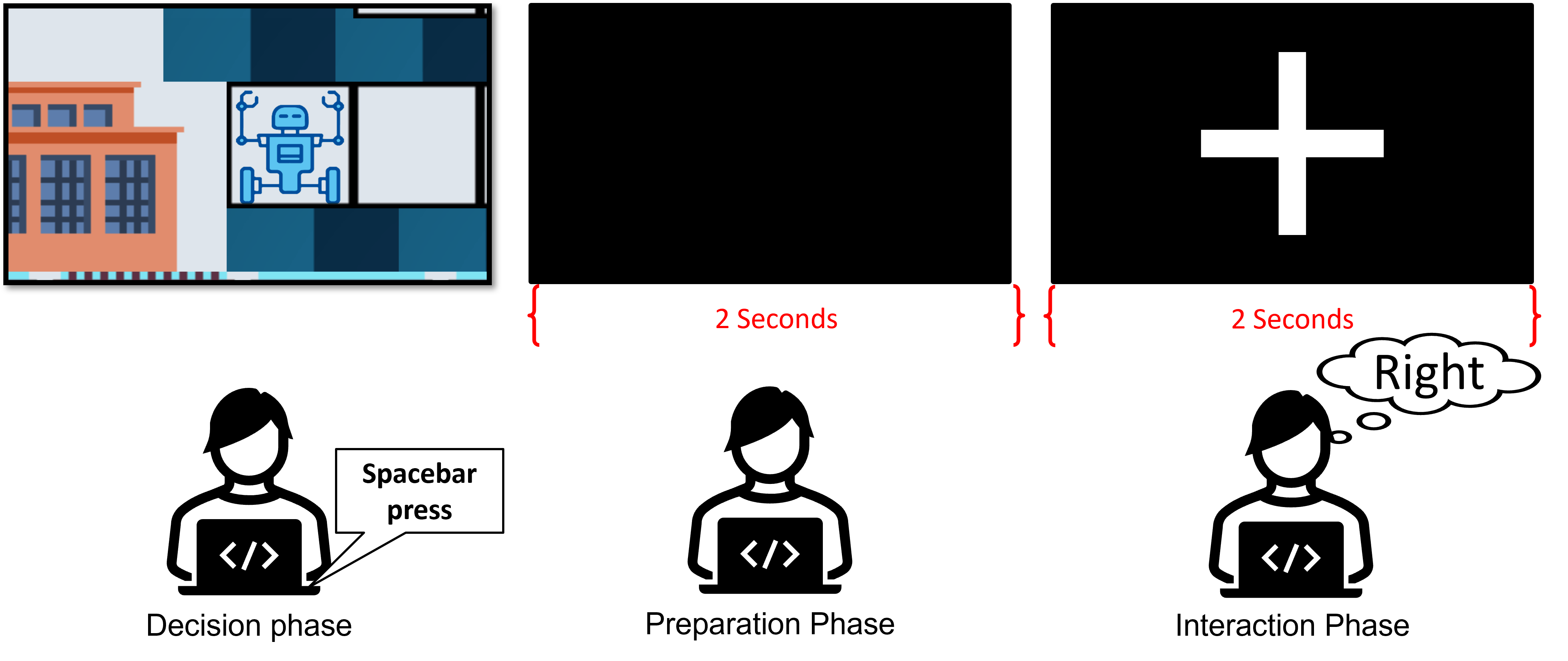}}
\caption{
    Protocol for collecting imagined speech data~\cite{mohamed2024speech}.
    A spacebar press triggers the interaction during overt speech, 
    while imagined speech interactions occur in separate sessions.
}
\label{fig:interactionRobotGame}
\end{figure}

EEG data was recorded from participants 
while they interacted with a virtual robot using both overt (spoken aloud) and covert (imagined) speech.
The participants viewed a bird's-eye view of the robot positioned at the center of the screen. 
The robot was controlled using five different command words:
\textsc{Left}, \textsc{Right}, \textsc{Up}, \textsc{Pick}, and \textsc{Push}.
The experiment was divided into four sessions: 
two for overt speech and two for covert speech. 
Each session consisted of eight levels. 
In each level, each of the five command words was performed five times in random order, 
resulting in 200 trials per session (8 levels $\times$ 5 words $\times$ 5 repetitions).
The sessions alternated between overt and covert conditions, to avoid recording data in large blocks, 
which could have led to classifying random brain states instead of actual cognitive processes.
Fig.~\ref{fig:interactionRobotGame} illustrates the experimental paradigm. 

\subsubsection{Procedure}
To allow participants to navigate the game at their own pace, 
we used the spacebar as the game controller (decision phase). 
Pressing that key caused a black screen to appear for two seconds, 
during which participants focused on which command they had to think (preparation phase), 
and when a white cross appeared in the middle of the screen for two additional seconds, 
as a cue, participants would produce the imagined speech (interaction phase).
A two-second rest period followed, and the process was then repeated. 
In each level, the robot randomly presented the five action words. 
In the overt session, participants spoke the command aloud. 
In the covert session, they silently repeated the word in their mind without moving their mouth or muscles. 
Each command word was repeated 80 times in both the covert and overt speech sessions, 
resulting in 400 repetitions for each condition and a total of 800 repetitions overall.

\subsubsection{Data collection}
EEG data were collected in a controlled environment, 
with participants seated in comfortable armchairs to minimize unnecessary movements. 
EEG signals were recorded using a wireless 64-channel EEG system (Brain Products Live Amp 64) 
at a sampling rate of 500\,Hz, following the 10-20 international EEG system.
The robotic experimental paradigm was executed on the same Windows PC used for EEG recording, 
allowing for the synchronization of game events (such as spacebar presses and fixation crosses) with the EEG data. 
The recorded data were stored on the PC for subsequent offline analysis.

\subsection{Data preprocessing and feature extraction}

The EEG data were stored separately for each participant 
and individually preprocessed for each session. 
A 50\,Hz notch filter was first applied to remove powerline interference. 
Subsequently, a 4th-order Butterworth Finite Impulse Response (FIR) filter 
with a passband of 0.5 to 80\,Hz was used to filter the data.
To eliminate artifacts related to eye blinks, muscles, and movement-related activities, 
independent component analysis (ICA)~\cite{iriarte2003independent} 
was applied to the filtered EEG data.
The EEG signal was then segmented into two-second epochs corresponding to the speech production phase, 
with a 100\,ms pre-stimulus window for baseline correction.
The mean signal from this 100\,ms pre-stimulus window was subtracted from the corresponding epoch
to minimize baseline drift and improve signal consistency.
Finally, individual trials were labeled as covert or overt speech for modeling.

The EEG data during speech reflect the brain's neural response to motor articulation and cognitive processes. 
To extract meaningful features from EEG signals for the classification of overt and imagined speech, 
we employed the \textit{Hilbert Envelope} (ENV) 
and \textit{Temporal Fine Structure} (TFS)~\cite{hilbert1989grundzuge,he2016praat} as key representations of EEG data. 
These features capture amplitude variations (using ENV, which are essential for speech-related activity) 
and phase information (using TFS, which is crucial for modeling neural dynamics during speech).
Compared to alternative features~\cite{saeidi2021neural} such as wavelet-based features~\cite{biswas2022wavelet,hernandez2021toward}, common spatial patterns~\cite{wang2013analysis}, and Riemannian geometry-based features~\cite{bakhshali2020eeg}, these representations provide enhanced capability for detecting subtle patterns in brain activity associated with speech tasks.

The Hilbert Envelope is the magnitude of an analytic signal $\tilde{a}(t)$, 
in our case EEG, and represents the slow-varying amplitude of that signal over time.
It is computed as:
\begin{equation}
    \text{ENV}(a(t)) = ||\tilde{a}(t)|| = \sqrt{a(t)^2 + \mathcal{H}\{a(t)\}^2}
\end{equation}
where the original EEG signal $a(t)$ acts as the real part, and the Hilbert-transformed signal $\mathcal{H}\{a(t)\}$ acts as the imaginary part. $\mathcal{H}\{\cdot\}$ denotes the Hilbert transform operation.

The Hilbert transform of a signal in the time domain is defined as the convolution of $a(t)$ with $\frac{1}{\pi t}$, which can be expressed as:
\begin{equation}
    \hat{a}(t) = \mathcal{H}\{a(t)\} = \frac{1}{\pi t} \ast a(t) = \frac{1}{\pi} \int_{-\infty}^{\infty} \frac{a(\tau)}{t - \tau} d\tau
\end{equation}
This envelope captures the amplitude modulation of the EEG signal, 
reflecting the overall energy changes that occur during speech production.

The Temporal Fine Structure, on the other hand, 
captures the rapid oscillations within the signal and is related to the phase information of the analytic signal. 
It is computed by normalizing the original EEG signal with its envelope:
\begin{equation}
    \text{TFS}(a(t)) = \frac{{a}(t)}{\text{ENV}(a(t))}
\end{equation}
The fine temporal details of the EEG signal are said to carry information about the neural dynamics underlying speech~\cite{moon2014temporal,ni2023eeg}.

The ENV and TFS features are extracted from the EEG data for each trial. 
Each EEG segment has an original dimension of 1000$\times$64.
After applying the ENV and TFS, each EEG segment is transformed into two feature segments of the same dimension. 
These segments are then concatenated horizontally, resulting in a new dimension of 1000$\times$128. 

\subsection{Dataset splits and model description}

We train subject-specific models using LSTM, Gated Recurrent Unit (GRU), BiLSTM, and Bidirectional GRU (BiGRU) architectures to classify EEG data. 
Each subject provided 400 EEG segments for covert speech and 400 segments for overt speech. 
We rely on transfer learning, where models are initially trained on overt speech EEG data. 
The BiLSTM model, which achieved the best performance, is then used for transfer learning. 
The LSTM layers of the BiLSTM model are frozen, 
and the fully connected layer is fine-tuned using 15\%, 20\%, 25\% and 30\% of the EEG features of covert speech
to create a covert speech classification model.

The BiLSTM model begins with a sequence input layer,
accepting an input size of $(B, 1000, 128)$, where $B$ represents the batch size (we used $B=32$), 
$1000$ is the number of time steps, and $128$ is the number of features. 
The model is trained with Adam optimizer using a learning rate $\eta=0.0001$ and decay rates $\beta_1=0.9, \beta_2=0.999$.
The loss function is categorical entropy.
Table~\ref{tab:model} shows the details of the model architecture. 

\begin{table}[!ht]
\centering
\caption{Model configuration of BiLSTM model.}
\label{tab:model}
\begin{tabular}{lll}
\toprule
\textbf{Layer Type} & \textbf{Learnable hyperparams} & \textbf{Remarks} \\
\midrule
\text{Input} & \text{None} & \text{Shape: (B, 1000, 128)} \\
\midrule
\text{BiLSTM 1} & \text{Input weights: 4$\times$128$\times$512} & \text{Hidden units: 512} \\
& \text{Recurrent weights: 4$\times$512$\times$512} & \text{State fn: tanh} \\
& \text{Bias: 4$\times$512} & \text{Gate fn: sigmoid} \\
& & \text{Dropout rate: 0.3} \\
\midrule
\text{BiLSTM 2} & \text{Input weights: 4$\times$1024$\times$256} & \text{Hidden units: 256} \\
& \text{Recurrent weights: 4$\times$256$\times$256} & \text{State fn: tanh} \\
& \text{Bias: 4$\times$256} & \text{Gate fn: sigmoid} \\
& & \text{Dropout rate: 0.2} \\
\midrule
\text{Fully Conn.} & \text{Weights: 256$\times$5} & \text{Size: 5} \\
& & \text{Bias: 5$\times$1} \\
\midrule
\text{Output} & \text{None} & \text{Softmax. Size: 5} \\
\bottomrule
\end{tabular}%
\label{tab:bi-lstm-model-config}
\end{table}

\section{Results and discussion}
\label{sec:Results a dn Discussions}

\subsection{Covert-Overt EEG decoding}

\begin{figure}[!ht]
    \centering
    \includegraphics[width=0.8\linewidth]{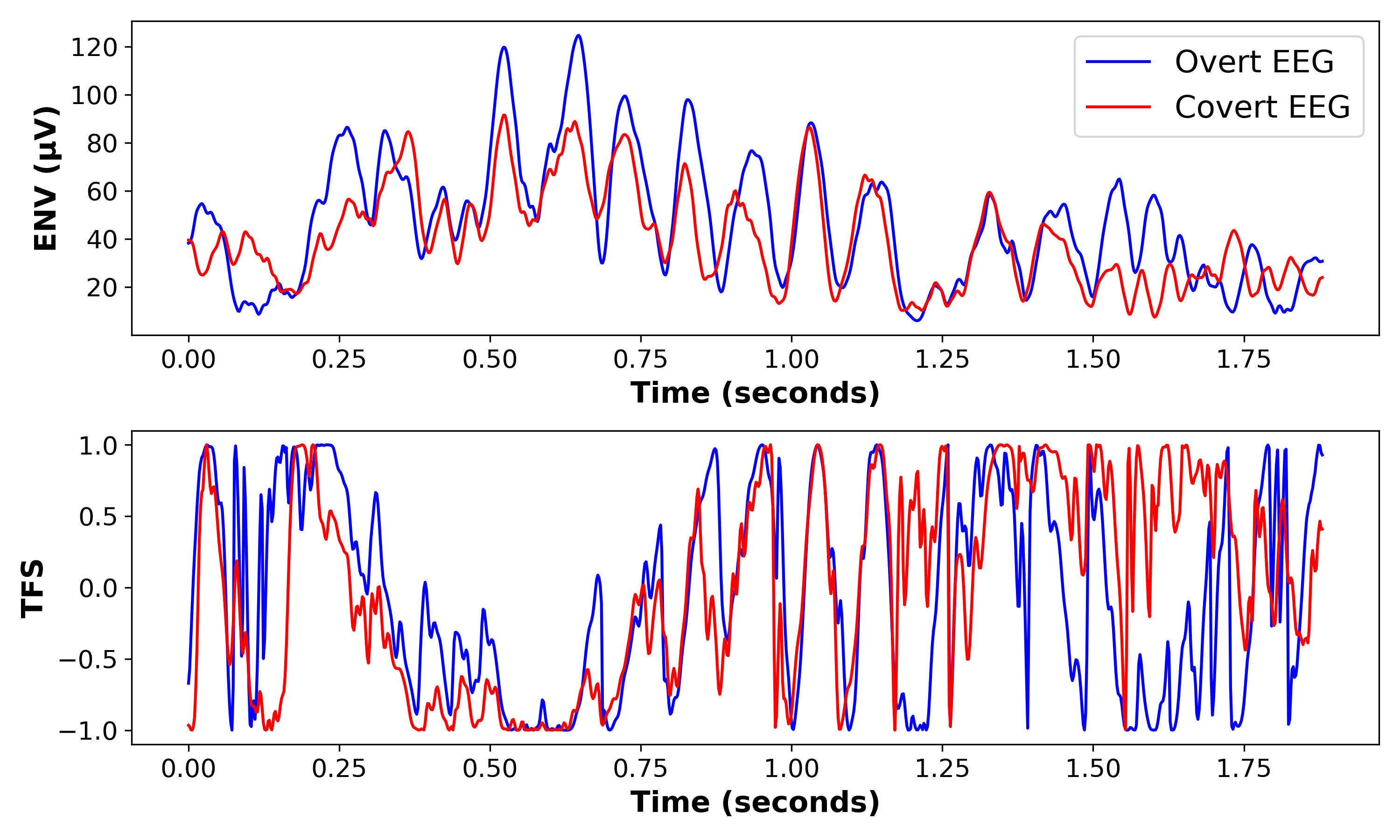}
    \caption{ENV and TFS of the speech command \textsc{Right}.}
    \label{fig:ExampleEyeBlinks}
\end{figure}

Previous work~\cite{watanabe2020synchronization, zhang2024revealing,oppenheim2010motor} revealed spatio-temporal dynamics and waveform synchronization between covert and overt speech EEG data. This foundation guided our feature extraction, using ENV and TFS to analyze such a waveform synchronization. Fig.~\ref{fig:ExampleEyeBlinks} shows the average Hilbert envelope and temporal fine structure for covert and overt speech commands of participant 10 (command: \textsc{Right}). The ENV and TFS of the EEG data for overt and covert speech show structural similarities. The max. correlation coefficients between the EEG envelopes of covert and overt speech for commands are 0.5234 for \textsc{Up}, 0.6412 for \textsc{Left}, 0.6607 for \textsc{Right}, 0.4791 for \textsc{Pick}, and 0.5624 for \textsc{Push}, indicating varying degrees of similarity in the envelopes of amplitude of the EEG signals. 

\subsection{Classification results}

The ENV and TFS features are combined as input to the models 
to learn the underlying EEG patterns during different speech tasks.
Four different subject-specific classification models 
were trained on the overt speech EEG data using a five-fold cross-validation approach. 
Among these, the BiLSTM model (average accuracy of 86.44\%) outperformed the others, 
making it the preferred model for transfer learning to covert EEG classification.
The BiLSTM's superior performance is attributed to its ability to process EEG data bidirectionally, effectively capturing both forward and backward temporal dependencies while leveraging the non-stationary and dynamic statistical characteristics of EEG signals to decode speech-related neural activity.
Table~\ref{tab:results} summarizes the results.

\begin{table}[!ht]
\centering
\caption{
Classification accuracy results, in percentage.
For each participant, the best result is highlighted in bold face.
}
\label{tab:results}
\begin{tabular}{*6c}
\toprule
    & \multicolumn{4}{c}{\textbf{Overt}} & \textbf{Covert} \\ 
    \cmidrule(rl){2-5} 
 \textbf{User} & \textbf{LSTM} & \textbf{GRU} & \textbf{BiLSTM} & \textbf{BiGRU} & \textbf{BiLSTM} \\ 
 \midrule
P1   & 70.37 & 63.78 & \best{75.15} & 64.80 & 58.92 \\ 
P2   & 66.34 & 63.21 & \best{71.80} & 62.39 & 58.33 \\ 
P3   & \best{87.00} & 83.50 & 86.30 & 84.50 & 86.21 \\ 
P4   & 88.68 & 85.77 & \best{90.65} & 86.28 & 85.90 \\ 
P5   & \best{85.00} & 74.96 & 84.33 & 73.90 & 85.72 \\ 
P6   & 91.00 & 90.32 & \best{93.07} & 91.17 & 86.20 \\ 
P7   & 94.11 & 93.43 & \best{96.53} & 92.47 & 89.74 \\ 
P8   & 84.40 & 80.50 & \best{88.46} & 81.69 & 79.17 \\ 
P9   & \best{86.00} & 80.74 & 85.22 & 79.80 & 73.36 \\ 
P10  & \best{95.32} & 95.00 & 95.00 & 94.37 & 90.81 \\ 
P11  & 90.07 & 91.84 & \best{92.38} & 90.90 & 86.55 \\ 
P12  & 86.45 & 84.00 & \best{90.49} & 85.53 & 83.68 \\ 
P13  & 71.28 & 70.03 & \best{76.82} & 71.00 & 67.34 \\ 
P14  & 79.95 & 83.40 & \best{84.00} & 82.30 & 85.60 \\ 
P15  & 75.71 & 77.50 & \best{80.86} & 78.60 & 68.59 \\ 
\midrule
\textbf{Avg. Accuracy (\%)} & 83.99 & 81.46 & \best{86.44} & 81.51 & 79.82 \\ 
\bottomrule
\end{tabular}%
\end{table}

\begin{figure}[!ht]
    \centering
    \includegraphics[width=0.7\linewidth]{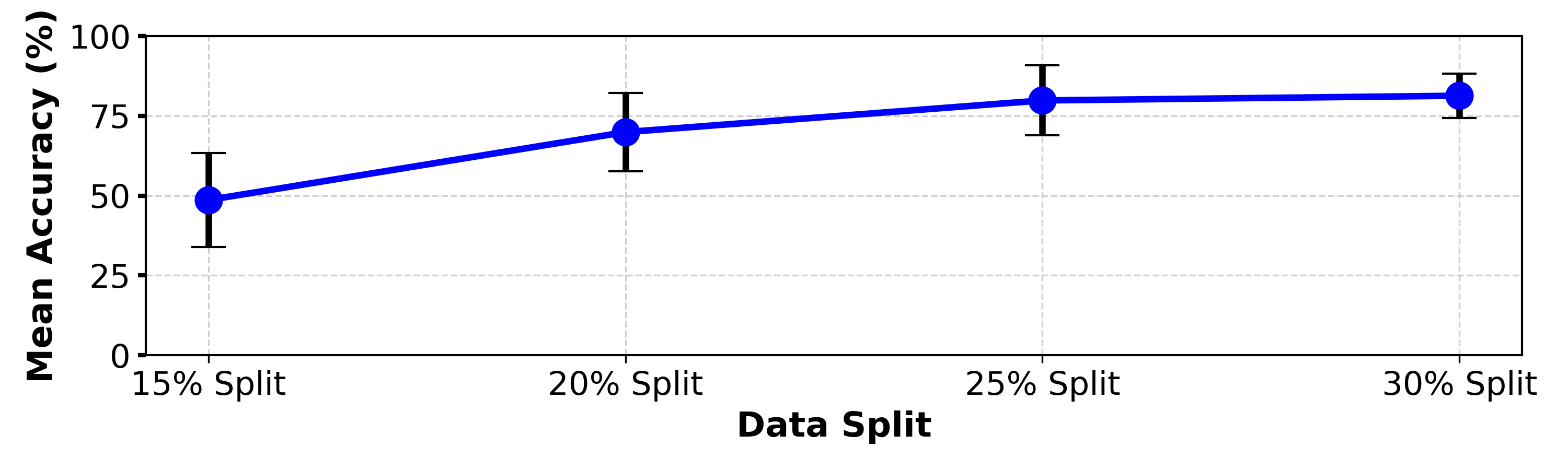}  
    \caption{BiLSTM model performance with varying data splits of covert EEG data. Error bars denote SDs.}
    \label{fig:Covert}
\end{figure}

For covert speech classification, the trained BiLSTM model was used with the LSTM layer weights frozen. 
The fully connected layers were re-trained using 15\%, 20\%, 25\%, and 30\% of the covert EEG data. 
To compare the classification accuracy of the transferred covert and standard overt speech models, 
20\% of the unknown covert EEG data were used to test the accuracy of the transferred covert model, 
as the standard overt classifier was trained and tested with a 80:20 split. 
Classification results are illustrated in Fig.~\ref{fig:Covert},
showing that with 25\% and 30\% of the covert training data, 
the classifier performance did not deviate significantly ($p<.05$, paired $t$-test with Bonferroni correction). 
Furthermore, the performance of the transferred model becomes more consistent 
and less variable when exposed to more training examples. 

We also trained the same models shown in Table~\ref{tab:results} from scratch,
specifically for covert speech EEG classification.
Table~\ref{tab:covert_results} summarizes the results, 
demonstrating that transfer learning significantly improves performance.
For example, while the BiLSTM model was also the best performer when trained from scratch,
it was nevertheless 20 points below the BiLSTM trained with transfer learning.

\begin{table}[!ht]
\centering
\caption{
    Covert speech EEG classification results (Accuracy $\pm$ StDev) without transfer learning.
}
\label{tab:covert_results}
\begin{tabular}{*5c}
\toprule
    & \textbf{LSTM} & \textbf{GRU} & \textbf{BiLSTM} & \textbf{BiGRU} \\ 
\midrule
\textbf{Avg. Accuracy (\%)} & 65.70 \(\pm\) 3.39 & 65.0 \(\pm\) 3.68 & 66.94 \(\pm\) 4.65 & 65.83 \(\pm\) 4.72 \\ 
\bottomrule
\end{tabular}%
\end{table}

As shown in Table~\ref{tab:comparison}, 
limited research has been conducted on the transfer of BCI models of overt speech to covert speech.
Our proposed implementation outperforms previous methods.
This was possible thanks to the waveform synchronization between overt and covert speech. 
Our work addresses the limitations of covert speech recording, 
including the challenges of data collection and reliability (e.g., mislabeling, variability in speech onset, offset, and speed). 

\begin{table}[!ht]
\caption{Comparison against previous work.}
\begin{center}
    \begin{tabular}{*5c}
    \toprule
    \textbf{Ref.} & \textbf{No. commands} & \textbf{Data} & \textbf{Classification acc.} \\
    \midrule
    \cite{lee2020eeg} & 12 & EEG   & 16.20\% \\
    \cite{komeiji2024feasibility} & 8  & ECoG  & 46.60\% \\
    \cite{cooney2019optimizing} & 5  & EEG   & 35.68\% \\
    \cite{rekrut2022improving} & 5  & EEG   & 61.78\% \\
    \cite{mohamed2024speech} & 5  & EEG   & 69.10\% \\
    \textbf{Ours} & 5 & EEG & \textbf{79.82\%} \\
    \bottomrule
    \end{tabular}
\label{tab:comparison}
\end{center}
\end{table}

\subsection{Limitations and Future Work}

Since EEG responses are highly subjective, the internal decoding of overt and covert word thoughts may vary between individuals. 
This variability poses a challenge in creating ``universal'' models. 
Future work could focus on analyzing waveform synchronization between participants
to address intersubject variability and explore speaker independence.

\section{Conclusion}
\label{sec:conclusion}

Our work emphasizes the importance of two key EEG features: Hilbert envelope and temporal fine structure, 
in decoding both overt and covert speech neural patterns. 
Our work also underscores the importance of transfer learning for accurately classifying covert speech EEG data.
We find that our BiLSTM model outperforms state-of-the-art techniques, 
representing significant advancements for training imagined speech BCIs.

\section*{Acknowledgments}
Research supported by the Horizon 2020 FET program of the European Union 
through the ERA-NET Cofund funding (grant CHIST-ERA-20-BCI-001)
and the Pathfinder program of the European Innovation Council (SYMBIOTIK project, grant 101071147). 
Rekrut’s work~\cite{rekrut2022improving} is supported by the German Federal Ministry of Education and Research (grants 01IS12050 and 01IS23073).

\balance
\bibliographystyle{IEEEbib}
\bibliography{refs}

\end{document}